\documentclass[letterpaper, 10 pt, conference]{ieeeconf}

\IEEEoverridecommandlockouts                         
\usepackage{bm}
\usepackage{amssymb}
\usepackage{threeparttable}
\usepackage{wrapfig}
\usepackage{amsmath}
\usepackage{wrapfig}
\usepackage{lipsum}
\usepackage{siunitx}
\usepackage{booktabs}
\usepackage{glossaries}
\usepackage{hyperref}
\usepackage{graphicx}
\usepackage{url}
\usepackage{cleveref}
\usepackage{cite}
\usepackage{times}
\usepackage[table]{xcolor}

\title{\LARGE \bf
iRoCo: Intuitive Robot Control From Anywhere Using a Smartwatch
}

\author{
Fabian C Weigend$^{1}$, 
Xiao Liu$^{1}$, 
Shubham Sonawani$^{1}$, 
Neelesh Kumar$^{2}$, \\
Venugopal Vasudevan$^{2}$ and
Heni Ben Amor$^{1}$
\thanks{$^{1}$Weigend, F, Liu, X, Sonawani, S \& Ben Amor, H are affiliated with SCAI, Arizona State University \texttt{$\lbrace$fweigend, xliu330, sdsonawa, hbenamor$\rbrace$@asu.edu}.
$^{2}$Kumar, N \& Vasudevan, V are affiliated with Corporate Functions-R\&D, Procter and Gamble
\texttt{$\lbrace$kumar.n.40,vasudevan.v$\rbrace$@pg.com}
}}

\begin{document}
\newacronym{gyro}{\ensuremath{\bm{\phi}}}{gyroscope measurements}
\newacronym{grav}{\ensuremath{\bm{\gamma}}}{gravity sensor}
\newacronym{lacc}{\ensuremath{\bm{\alpha}}}{linear acceleration sensor}
\newacronym{racc}{\ensuremath{\bm{\alpha}_\mathrm{raw}}}{raw acceleration}
\newacronym{swrot}{\ensuremath{\bm{\theta}}}{virtual rotation vector sensor}
\newacronym{swrot_calib}{\ensuremath{\bm{\theta}_c}}{calibration forward-facing direction}
\newacronym{pres}{\ensuremath{\rho}}{atmospheric pressure sensor}
\newacronym{rot_hip}{\ensuremath{\mathbf{q}_\mathrm{h}}}{hip rotation}
\newacronym{rot_larm}{\ensuremath{\mathbf{q}_\mathrm{l}}}{lower arm rotation}
\newacronym{rot_uarm}{\ensuremath{\mathbf{q}_\mathrm{u}}}{upper arm rotation}
\newacronym{rot_larm_r}{\ensuremath{\mathbf{q}^r_\mathrm{l}}}{relative lower arm rotation}
\newacronym{rot_uarm_r}{\ensuremath{\mathbf{q}^r_\mathrm{u}}}{relative upper arm rotation}
\newacronym{6drr}{6DRR}{six-dimensional rotation representation}

\maketitle
\thispagestyle{empty}
\pagestyle{empty}

\begin{abstract}
This paper introduces iRoCo (intuitive Robot Control) -- a framework for ubiquitous human-robot collaboration using a single smartwatch and smartphone. By integrating probabilistic differentiable filters, iRoCo optimizes a combination of precise robot control and unrestricted user movement from ubiquitous devices. 
We demonstrate and evaluate the effectiveness of iRoCo in practical teleoperation and drone piloting applications. Comparative analysis shows no significant difference between task performance with iRoCo and gold-standard control systems in teleoperation tasks. Additionally, iRoCo users complete drone piloting tasks 32\% faster than with a traditional remote control and report less frustration in a subjective load index questionnaire. Our findings strongly suggest that iRoCo is a promising new approach for intuitive robot control through smartwatches and smartphones from anywhere, at any time. The code is available at \mbox{\url{www.github.com/wearable-motion-capture}}
\end{abstract}

\section{Introduction}
Human motion tracking and state estimation are essential components of many robotics approaches, such as teleoperation~\cite{ajoudani2018progress}, exoskeleton control~\cite{8968007}, learning from demonstration~\cite{NIPS1996_68d13cf2}, and in particular human-robot collaboration~\cite{BenAmorNKKP2014}. A major challenge faced in these fields is the reliance on costly motion capture systems, e.g, OptiTrack ~\cite{nagymate_application_2018}, which are restricted to stationary setups and require elaborate calibration procedures. For more flexibility, non-optical motion capture systems utilizing Inertial Measurement Units (IMUs) have gained traction \cite{DESMARAIS2021103275,DBLP:journals/corr/MarcardRBP17}. However, most IMU-based methods still require precise placement of specialized units on the user's body and intricate calibration procedures~\cite{yi2021transpose,roetenberg2009xsens}.

To address these drawbacks and enable human pose estimation outside of the lab, recent works have investigated the potential of smart devices for IMU-based motion capture. These approaches promise truly ubiquitous human-robot collaboration, as motion capture through smart devices is widely accessible and intuitive for untrained users~\cite{villani2020humans, weigend2023anytime}. Studies have specifically examined the use of a single smartwatch for motion capture~\cite{shen_i_2016, wei2021real, liu2022real}, along with the development of human-robot control interfaces based on wearable technology~\cite{villani2017interacting, villani2020humans, weigend2023anytime}. However, existing works in smartwatch motion capture impose constraints on user movements and orientation, thereby reducing their utility and practicality~\cite{weigend2023anytime,wei2021real,shen_i_2016}. 

% motion capture from a single smartwatch with sufficient accuracy for robot control remains challenging. Pr

\begin{figure}[t!]
\centering
\includegraphics[width=\linewidth]{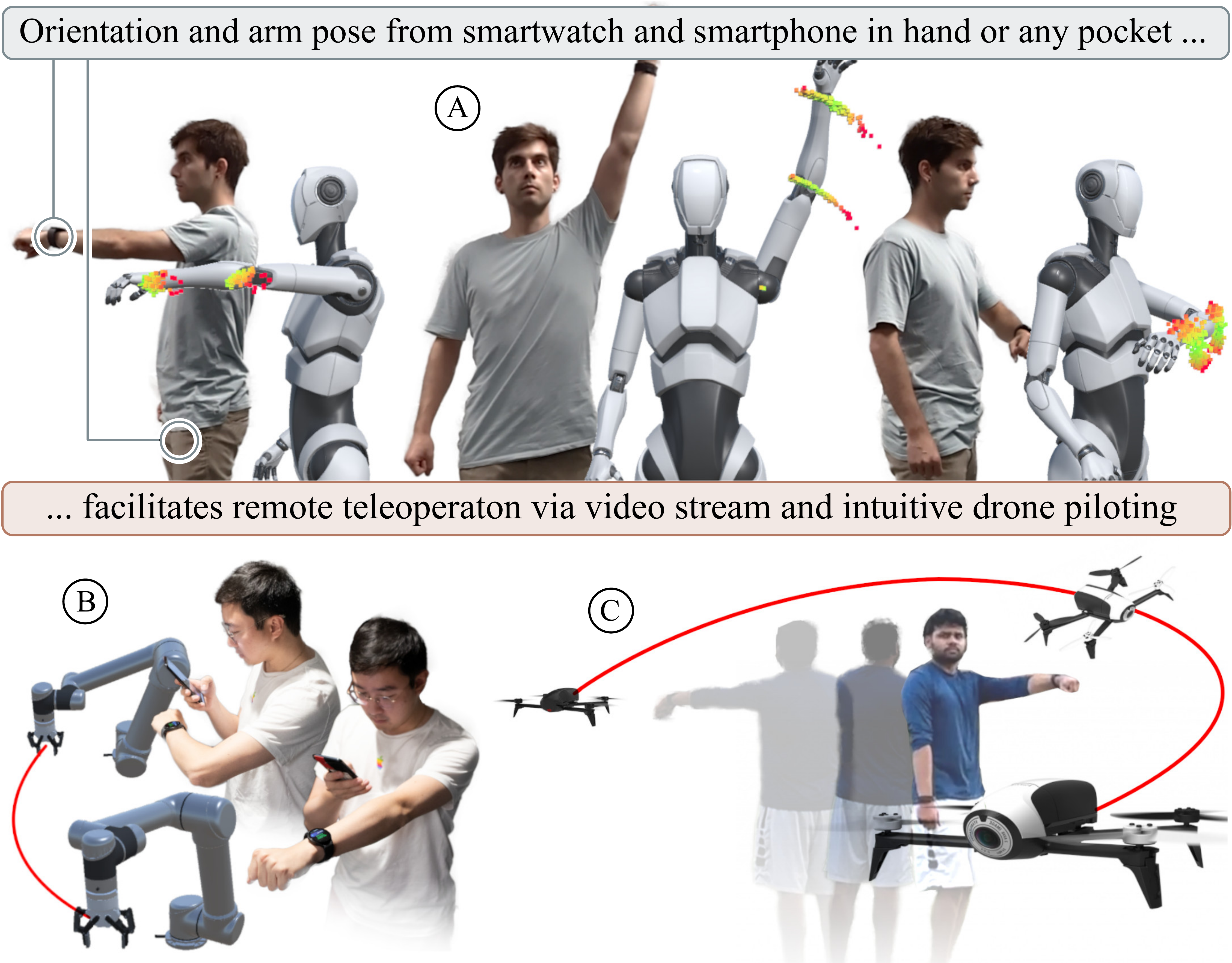}
\caption{{\bf A)} We present an \textbf{i}ntuitive \textbf{Ro}bot \textbf{Co}ntrol (iRoCo) framework, which achieves robust body orientation and arm pose estimations from a smartwatch and a smartphone. {\bf B)} The system allows teleoperation through video streaming to a smartphone, enabling tasks such as pick-and-place operations. {\bf C)} We demonstrate iRoCo's potential for intuitive drone piloting.}
\label{fig:overview}
\end{figure} 

This paper expands upon previous work by augmenting the sensor data from a smartwatch with a connected smartphone. Using the combined data, we leverage advancements in neural state estimation techniques, specifically differentiable recursive filters~\cite{kloss2021train,liu2023enhancing,liu2023alpha},
to allow users to move freely in their environment, thereby addressing limitations of previous approaches for smartwatch motion capture. We employ a differentiable Bayesian state estimator, augment it with a tailored control modality and present the entire system as {\bf i}ntuitive {\bf Ro}bot {\bf Co}ntrol (iRoCo). As illustrated in~\Cref{fig:overview}, iRoCo optimizes the balance between unrestricted movements and reliable pose estimations for various human-robot interaction tasks. %Users are not limited to fixed body orientations and motion capture remains ubiquitous. 
We demonstrate the advantages of iRoCo by evaluating the user experience and comparing it to gold-standard baselines in teleoperation and drone piloting tasks. We summarize our contributions as: 

\begin{itemize}
    \item Introducing iRoCo, an intuitive robot control interface enabling robot control anytime and anywhere from a single smartwatch and smartphone.
    \item iRoCo integrates a differentiable filter algorithm for refined human pose estimation with uncertainty. 
    \item We design a tailored control modality for ubiquitous and intuitive teleoperation by a human operator.
    \item We demonstrate the advantages of iRoCo in two applications: teleoperated pick-and-place tasks and drone piloting.
\end{itemize}

\section{Related Work}
Smart device-based human-robot systems provide a wide range of possibilities to enhance the flexibility and robustness of human-robot interactions. These include navigating mobile robots using a single smartwatch~\cite{villani2017interacting}, automated monitoring of vital body functions through smartwatches~\cite{lee2015smartwatch, wicaksono2022towards}, and the coordination of multi-robot systems~\cite{villani2020humans}. The utilization of wearable devices has demonstrated the potential to alleviate user fatigue and reduce mental workload in various robot control tasks~\cite{villani2020humans}. However, numerous applications heavily rely on accurate gesture recognition by human users. Consequently, these systems suffer from reduced precision, despite their intuitive nature.

Aware of these challenges, the works of~\cite{shen_i_2016, wei2021real, liu2022real} proposed approaches to make more accurate arm pose predictions from smartwatch sensor data. In previous work, we presented approaches to utilize such arm pose predictions for teleoperation and robot control through voice commands~\cite{weigend2023anytime}. While these systems are promising for applications in robotics, they impose calibration procedures and fixed body-forward facing constraints on the user~\cite{shen_i_2016, wei2021real}. Or, they limit inference to the same environment where the model was trained \cite{liu2022real}.

%One notable exception is the recent work in~\cite{imuPoser}, where they propose a method for full human pose tracking using one or multiple smart devices using LSTM model, regardless of the human's orientation. 
% Although this approach, referred to as ``imuPoser", is capable of providing a best-guess pose consistently, it still does not attain the desired precision necessary for precise robot control.

To address the limitations associated with human arm pose estimation from smart devices, we propose the utilization of differentiable filters (DFs)~\cite{kloss2021train,liu2023enhancing,liu2023alpha}. DFs are a subclass of algorithms derived from the Deep State-Space model~\cite{NEURIPS2018_5cf68969, klushyn2021latent, kloss2021train} and offer a solution to combine the principles of Bayesian recursive filtering while learning state transition and measurement models from data. Previous research\cite{kloss2021train} has demonstrated that DFs can effectively learn noise profiles through end-to-end training, and their efficacy has been evidenced in various real-world applications\cite{wagstaff2022self, lee2020multimodal, liu2023learning}.  In our study, we employ the Differentiable Ensemble Kalman Filter (DEnKF)~\cite{liu2023enhancing} as the stateful model for accurate arm pose estimation and uncertainty measurement. By utilizing both a smartwatch and a smartphone, our system enables users to have greater freedom of movement. The DEnKF is leveraged as the inference model, ensuring stable and precise pose estimations throughout our experiments.

\section{Methodology}

%As illustrated in \Cref{fig:method}, 
This section describes data collection, defines states, observations, and the implementation details of our DEKnF for human pose state estimations.

\subsection{Data Collection, Observation, and State}\label{sec:data}
\begin{figure}[t]
\centering
\includegraphics[width=\linewidth]{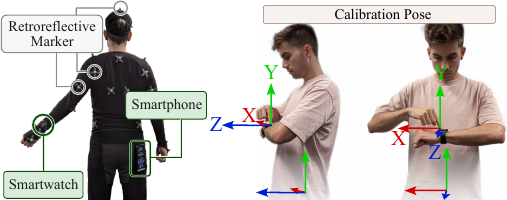}
\caption{{\bf Left}: Data collection to record ground truth OptiTrack poses together with smartwatch and smartphone sensor data. {\bf Right}: 
Starting our app on the watch calibrates the watch and phone orientation. The start pose sets the local coordinate system and defines the forward direction.}
%The DEnKF model structure. The stochastic Transition Model forwards the ensemble ${\bf X}_{t-1}$ one step in $t$. The Sensor Model projects raw observations to the observation space, such that the KF Update step corrects the state ensemble at ${\bf X}_{t}$.}
\label{fig:data_collection}
\end{figure}

For data collection, we developed smartwatch (Wear OS) and smartphone (Android) apps to stream sensor data, and made use of the optical motion capture system Optitrack  \cite{nagymate_application_2018} for capturing ground truth. %The apps for the smartwatch and smartphone stream sensor data to a remote machine. 
As depicted on the left in \Cref{fig:data_collection}, the human subjects wore a 25-marker Optitrack suit, a smartwatch on their left wrist, and kept a smartphone in their pocket. After starting the data recording with watch, phone, and Optitrack, we asked subjects to perform random arm motions and encouraged them to vary their body orientation or move around within the optically trackable area. Recorded data comprises of calibrated observation data from watch and phone together with ground-truth state data from OptiTrack. We define calibrated observations and states in the following.
 
\textbf{Observation:} The raw observation $\bf{y}$ consists of the values \mbox{
${\bf y} = [ \Delta t, \pmb{\theta}_{\mathrm{sw}}, 
    \pmb{v}, 
    \pmb{\alpha},
    \pmb{\gamma},
    \pmb{\phi},
    \rho,
    \mathbf{r}_\mathrm{h}
]^\top$}, with \mbox{${\bf y} \in \mathbb{R}^{22}$}. The following values are measured by the smartwatch: The \mbox{\gls{swrot}} by Wear OS provides a global orientation with respect to the magnetic North and gravity vector. We transform the provided quaternion into the form of a continuous \gls{6drr}, which is well-suited for training neural networks~\cite{zhou_continuity_2019}. Therefore, \mbox{$\gls{swrot}_\mathrm{sw} \in \mathbb{R}^6$}. The measurements \mbox{$\pmb{\alpha},\pmb{\gamma},\pmb{\phi} \in \mathbb{R}^3$} are the IMU readings. Further, we integrate linear acceleration measurements between two observations over $\Delta t$ and denote them as velocities $\pmb{v} \in \mathbb{R}^3$. The value $\rho$ is the atmospheric pressure sensor. Finally, ${\bf r}_\mathrm{h}$ is measured by the smartphone: This measurement is the rotation around the calibrated up-axis of the virtual rotation vector sensor provided by Android. To turn this rotation into a continuous rotation representation, we estimate ${\bf r}_\mathrm{h}$ as the sine and cosine of the rotation angle. Therefore, ${\bf r}_\mathrm{h} \in \mathbb{R}^2$.

% \begin{wrapfigure}{r}{0.5\linewidth}
%     \includegraphics[width=\linewidth]{figures/f_calib.pdf}
%     \caption{Starting our app on the watch calibrates the watch and phone orientation. The start pose sets the local coordinate system and defines the forward direction.}
%     \label{fig:calib}
% \vspace{-0.1in}
% \end{wrapfigure}

\textbf{Calibration:} We require a calibration to bring the virtual rotation sensors of watch and phone into the reference frame of the human body, and to define forward and up directions. %as well as the atmospheric pressure sensor data. 
For a seamless calibration procedure, we propose the start pose as depicted in \Cref{fig:data_collection}. When the user starts the app on the smartwatch, they hold their lower arm parallel to the chest and hip. We then record subsequent watch orientation measurements $\pmb{\theta}_\mathrm{sw}$ and phone orientation measurements ${\bf r}_\mathrm{h}$ relative to their orientations in this start pose. %Because the user holds the lower arm parallel to the hip, this pose brings watch, phone, and body orientation measurements into the same reference frame. 
Further, because our simulations run in the Unity engine, we transform them into a right-handed coordinate system with the Y-axis up and Z-axis forward. In addition to the orientations, we also calibrate atmospheric pressure $\rho$ by recording subsequent measurements relative to the initial measurement in start pose. Therefore, calibrated $\pmb{\theta}_\mathrm{sw}$, ${\bf r}_\mathrm{h}$, and $\rho$ measurements are recorded relative to their initial values in the initial calibration pose when the user started streaming.

% The user carries a smartphone. We also use the start position in \Cref{fig:data_collection} to calibrate the phone with the initial body-forward facing direction and then track changes using the orientation sensor from the phone. More specifically, we estimate an offset rotation from the phone to the forward direction, bringing it into the same global reference frame as the watch. We collect the body orientation as ${\bf r}_h$.

\textbf{State:} Given our calibrated observations, we define state as a vector holding the arm pose and forward-facing direction of the human and their velocities. %The ground truth values were recorded with the research-grade optical motion capture system OptiTrack. We record data from participants who wore a 25-marker-upper-body suit along with the smartwatch on their left wrist and carried a smartphone in their pocket (see \Cref{fig:data_collection}).
We utilize \mbox{\gls{rot_uarm}} and \mbox{\gls{rot_larm}} together with their velocities $\dot{\mathbf{q}_\mathrm{u}}$ and $\dot{\mathbf{q}_\mathrm{l}}$ 
in the continuous \gls{6drr}. %with calculated velocities in \gls{6drr} space, . 
Further, we facilitate the body-forward facing direction as the sine and cosine of the rotation around the body up-axis as ${\bf q}_\mathrm{h} \in \mathbb{R}^2$ and the angular velocity as Euler angles $\dot{{\bf q}_\mathrm{h}} \in \mathbb{R}$. The entire state is denoted as \mbox{$\mathbf{x} = [\text{\gls{rot_uarm}}, \text{\gls{rot_larm}}, \mathbf{q}_\mathrm{h}, \dot{\mathbf{q}_\mathrm{u}}, \dot{{\bf q}_\mathrm{l}}, \dot{{\bf q}_\mathrm{h}}]^\top$}, where $\mathbf{x} \in \mathbb{R}^{27}$.

\subsection{Differentiable Ensemble Kalman Filter}
\label{sec:DEnKF}

\begin{figure}[t!]
\centering
\includegraphics[width=\linewidth]{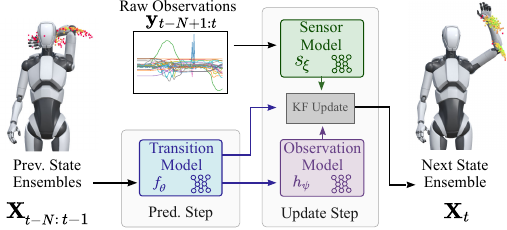}
\caption{The DEnKF model structure. In the Prediction Step, the stochastic Transition Model forwards the ensemble ${\bf X}_{{t-N}:{t-1}}$ one step in time (${\bf \tilde{X}}_t$). In the following Update Step, the Sensor Model projects raw observations to the observation space and the Observation Model projects ${\bf \tilde{X}}_t$ to the observation space, such that the KF Update corrects  ${\bf \tilde{X}}_t$ and we obtain ${\bf X}_t$.}
\label{fig:method}
\end{figure} 

To model the temporal transition between human pose states $\mathbf{x}$ and to map raw observations $\mathbf{y}$ to the state space, we utilize a Differentiable Ensemble Kalman Filter (DEnKF)~\cite{liu2023enhancing}. This state estimation approach enables us to learn and infer the dynamics of the human pose over time, while efficiently incorporating and processing the observed data. As depicted in \Cref{fig:method}, the DEnKF keeps the core algorithmic steps of an Ensemble Kalman Filter (EnKF)~\cite{evensen2003ensemble} while leveraging the capabilities of stochastic neural networks (SNNs)~\cite{gal2016dropout}. There are two steps in DEnKF, the \mbox{\emph{Prediction Step}} propagates the state one step further in time, and the \mbox{\emph{Update Step}} corrects the state based on newly collected observations. We define the learned observation for DEnKF as \mbox{$\tilde{{\bf y}} = [\text{\gls{rot_uarm}}, \text{\gls{rot_larm}}, \mathbf{q}_\mathrm{h}, \dot{\mathbf{q}_\mathrm{u}}, \dot{{\bf q}_\mathrm{l}}, \dot{{\bf q}_\mathrm{h}}]^\top$.} 
Let ${\bf X}_{0:N}$ denote the states of $N$ steps in $t$ with a number of $E \in \mathbb{Z}^+$ ensemble members, we initialize the filtering process with ${\bf X}_{0:N} = [ {\bf x}^{1}_{0:N}, \dots, {\bf x}^{E}_{0:N}]$. 

\vspace{0.1in}
\textbf{Prediction Step}: This step is modeled using the Transition Model as shown in \Cref{fig:method}. We use a window of $N$ and we leverage the stochastic forward passes from a trained state transition model to update each ensemble member: 
    \begin{equation}
    \begin{aligned}\label{eq:1}
          {\bf \tilde{x}}^{i}_{t} & \thicksim  f_{\pmb {\theta}} ({\bf \tilde{x}}^{i}_{t}|{\bf x}^{i}_{t-N:t-1}),\  \forall i \in E.
    \end{aligned}
   \end{equation}
 Matrix ${\bf \tilde{X}}_{t} = [{\bf \tilde{x}}^{1}_{t}, \cdots, {\bf \tilde{x}}^{E}_{t}]$ holds the updated ensemble members which are propagated one step forward through the state space. Note that sampling from the transition model $f_{\pmb {\theta}}(\cdot)$ implicitly introduces a process noise.

\vspace{0.1in}
\textbf{Update Step}: Given the updated ensemble members ${\bf \tilde{X}}_{t}$, a nonlinear observation model $h_{\pmb {\psi}}(\cdot)$ is applied to transform the ensemble members from the state space to observation space. The observation model is realized via a neural network with weights $\pmb {\psi}$:
    \begin{align}
    \label{eq:2}
        {\bf H}_t {\bf \tilde{X}}_{t} &= \left[ h_{\pmb {\psi}}({\bf \tilde{x}}^1_{t}), \cdots, h_{\pmb {\psi}}({\bf \tilde{x}}^E_{t}) \right],\\
        \label{eq:3}
        {\bf H}_t {\bf A}_{t} &=  {\bf H}_t {\bf \tilde{X}}_{t} 
        - \left[\frac{1}{E} \sum_{i=1}^E h_{\pmb {\psi}}({\bf \tilde{x}}^i_{t}),
        \cdots,
        \frac{1}{E} \sum_{i=1}^E h_{\pmb {\psi}}({\bf \tilde{x}}^i_{t})\right]. \nonumber
    \end{align}
${\bf H}_t {\bf \tilde{X}}_{t}$ is the predicted observation, and ${\bf H}_t {\bf A}_{t}$ is the sample mean of the predicted observation at $t$. EnKF treats observations as random variables~\cite{evensen2003ensemble}.
% Hence, the ensemble can incorporate a measurement perturbed by a small stochastic noise thereby accurately reflecting the error covariance of the best state estimate~\cite{evensen2003ensemble}. 

As shown in \Cref{fig:method}, we incorporate a Sensor Model that can learn projections between the learned observation and raw observation space. To this end, we leverage the methodology of SNN to train a stochastic sensor model that takes N steps of the raw observation and predicts the current learned observation using $s_{\pmb {\xi}}(\cdot)$:
    \begin{equation}
    \begin{aligned}\label{eq:sensor}
          \tilde{{\bf y}}^{i}_t & \thicksim  s_{\pmb {\xi}} (\tilde{{\bf y}}^{i}_t|{\bf y}_{t-N+1:t}),\  \forall i \in E,\\
    \end{aligned}
   \end{equation}
where ${\bf y}_{t-N+1:t}$ represents the noisy observations. Sampling yields observations $\tilde{{\bf Y}}_t = [\tilde{{\bf y}}^{1}_t, \cdots, \tilde{{\bf y}}^{E}_t]$ and sample mean $\tilde{{\bf y}}_t = \frac{1}{E}\sum_{i=1}^E\tilde{{\bf y}}^i_t$. 

DEnFK then proceeds with the Kalman Filter Update (KF Update in \Cref{fig:method}). To this end, the innovation covariance ${\bf S}_t$ can then be calculated as:
    \begin{equation}
    \begin{aligned}\label{eq:4}
        {\bf S}_t &= \frac{1}{E-1}  ({\bf H}_t {\bf A}_t)  ({\bf H}_t {\bf A}_t)^T + r_{\pmb {\zeta}}(\tilde{{\bf y}_t}),
    \end{aligned}
    \end{equation}
where $r_{\pmb {\zeta}}(\cdot)$ is the measurement noise model implemented using MLP. We use the same way to model the observation noise as in~\cite{kloss2021train}, $r_{\pmb {\zeta}}(\cdot)$ takes a learned observation $\tilde{{\bf y}_t}$ in time $t$ and provides stochastic noise in the observation space by constructing the diagonal of the noise covariance matrix. The final estimate of the ensemble ${\bf X}_{t}$ can be obtained by performing the measurement update step:
    \begin{equation}
    \begin{aligned}\label{eq:5}
        {\bf A}_t = {\bf \tilde{X}}_{t} - \frac{1}{E}\sum_{i=1}^E&{\bf \tilde{x}}^i_{t},\ {\bf K}_t = \frac{1}{E-1} {\bf A}_t ({\bf H}_t {\bf A}_t)^T {\bf S}_t^{-1},\\
        {\bf X}_{t} &= {\bf \tilde{X}}_{t} + {\bf K}_t (\tilde{{\bf Y}}_t - {\bf H}_t {\bf \tilde{X}}_{t}),
    \end{aligned}
    \end{equation}
    % \begin{align}
    % \begin{split}\label{eq:5}
    %     {\bf A}_t = {\bf X}_{t} - \frac{1}{E}\sum_{i=1}^E{\bf x}^i_{t},\ {\bf K}_t = \frac{1}{E-1} {\bf A}_t ({\bf H}_t {\bf A}_t)^T {\bf S}_t^{-1}
    % \end{split}\\
    % % \begin{split}\label{eq:6}
    % %  {\bf K}_t &= \frac{1}{E-1} {\bf A}_t ({\bf H}_t {\bf A}_t)^T {\bf S}_t^{-1},
    % % \end{split}\\
    % \begin{split}\label{eq:7}
    % {\bf X}_{t|t} = {\bf X}_{t} + {\bf K}_t (\tilde{{\bf Y}}_t - {\bf H}_t {\bf X}_{t}),
    % \end{split}
    % \end{align}
where ${\bf K}_t$ is the Kalman gain. In inference, the ensemble mean ${\bf \bar{x}}_{t} = \frac{1}{E}\sum_{i=1}^E {\bf x}^i_{t}$ is used as the updated state. The neural network structures for all learnable modules are described in Table~\ref{tab:EnKF_module}, where the ReLU activation function is applied within all hidden layers and a liner activation function is used for the output layer in each sub-module.

\begin{table}[h!]
  \centering
  \caption{DEnKF learnable sub-modules.}
  \label{tab:EnKF_module}
  \scalebox{0.95}{
  \begin{tabular}{ll}
    \toprule
$f_{\pmb {\theta}}$: & 1$\times$SNN(256, ReLU), 1$\times$SNN(512, ReLU), 1$\times$fc(S, -)\\
$h_{\pmb {\psi}}$: & 2$\times$fc(32, ReLU), 2$\times$fc(64, ReLU), 1$\times$ fc(O, -)\\
$r_{\pmb {\zeta}}$: & 2$\times$fc(16, ReLU), 1$\times$fc(O, -)\\
$s_{\pmb {\xi}}$: & 2$\times$SNN(256, ReLU), 2$\times$SNN(64, ReLU), 1$\times$fc(O, -)\\
% \multirow{3}{1em}{$s_{\pmb {\xi}}$:} & conv(7$\times$7, 64, stride 2, ReLU), conv(3$\times$3, 32, stride 2, ReLU), \\
%  & conv(3$\times$3, 16, stride 2, ReLU), flatten(), 2$\times$SNN(64, ReLU),  \\
%  & 2$\times$SNN(32, ReLU), 1$\times$SNN(O, -)\\
    \bottomrule
\multicolumn{2}{l}{fc: fully connected, S, O: state and observation dimension.} \\
  \end{tabular}}
\end{table}

% \vspace{0.1in}

% \begin{figure}[t]
%     \centering
%     \includegraphics[width=\linewidth]{figures/preidction_examples.pdf}
%     \caption{Example predictions for motions in our test dataset. Namely, they are {\bf A)} cross chest, {\bf B)} wave, {\bf C)} walk figure eight, and {\bf D)} cross behind the head. Each ensemble member of the DEnKF algorithm is indicated as a cube, colored according to their distance from the mean. 
%     % The distributions allow for a measure of uncertainty. For example, they suggest a low confidence in the last pose.
%     }
%     \label{fig:pred_examples}
% \vspace{-0.2in}
% \end{figure}

% \vspace{0.1in}
\textbf{Training:} We train the entire framework in an end-to-end manner using a mean squared error (MSE) loss between the ground truth state $\hat{{\bf x}}_{t}$ and the estimated state ${\bf \bar{x}}_{t}$ at every timestep. We also supervise the intermediate modules via loss gradients $\mathcal{L}_{f_{\pmb {\theta}}}$ and $\mathcal{L}_{s_{\pmb {\xi}}}$. Given ground truth at time $t$, we apply the MSE loss gradient calculated between $\hat{{\bf x}}_{t}$ and the output of the state transition model to $f_{\pmb {\theta}}$ as in Eq.~\ref{eq:loss1}. We apply the intermediate loss gradients computed based on the ground truth observation $\hat{{\bf y}}_t$ and the output of the stochastic sensor model $\tilde{{\bf y}}_t$: 
    \begin{align}
    \label{eq:loss1}
    \mathcal{L}_{f_{\pmb {\theta}}} = \| f_{\pmb {\theta}}(\tilde{{\bf y}}_{t-N:t-1}) - \hat{{\bf x}}_{t}\|_2^2,\ \ 
        \mathcal{L}_{s_{\pmb {\xi}}} =\| \tilde{{\bf y}}_t -  \hat{{\bf y}}_t\|_2^2.
    \end{align}
All models in the experiments were trained for 50 epochs with batch size 256, a learning rate of $\eta = 10^{-4}$, and an ensemble size of 32. %We selected the model with the best performance on our test dataset. 

\subsection{Control Modality}
\label{subsec:control}

\begin{figure}[t]
    \centering
    \includegraphics[width=\linewidth]{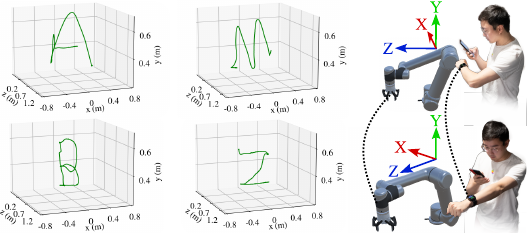}
    \caption{We propose a control modality where the body forward-facing direction together with the estimated wrist position on the sagittal plane determine the final end-effector position relative to the robot's base. This allows fine-grained control as demonstrated by writing the letters A, M, B, Z with the robot end-effector.}
    \label{fig:modality}
\end{figure}

We introduce a control modality to leverage human pose estimations for fine-grained robot control. As depicted in \Cref{fig:modality}, we utilize the predicted body forward-facing direction $\mathbf{q}_\mathrm{h}$ to set the local coordinate system for the user. The target robot end-effector position then is the projected wrist position on the sagittal plane (defined by the Z-axis and Y-axis). On the left in \Cref{fig:modality}, two human subjects used this modality to guide the end-effector of a Universal Robot 5 (UR5) within a $1.6\,\text{m} \times  0.6\,\text{m} \times 1\,\text{m}$ area to write letters. The trajectories of the end-effector were recorded with OptiTrack.
%The global rotation of the base and end-effector elevation and distance to the base. This control modality allows to effectively position the robot with arm pose and orientation.
% Two users demonstrate the precision of this modality in \Cref{fig:trajectories} by moving a Universal Robot 5 (UR5) end-effector equipped with retroreflective markers for motion capture within a $1.6\,\text{m} \times  0.6\,\text{m} \times 1\,\text{m}$ area used to write letters and numbers.
This completes our iRoCo system: the DEnKF algorithm to track the human pose from smartwatch and smartphone data together with the established modality for fine-grained robot control. %ng, we further demonstrate the effectiveness of iRoCo in use cases on real robots.

\section{Evaluation}

This section discusses the evaluation of the pose estimation accuracy of the DEnKF algorithm and outlines the composition of our training and test datasets. 

\subsection{Training and Test Datasets}
Training and test data was collected using the methodology described in \Cref{sec:data} and was approved by the institutional review board (IRB) of ASU under the ID STUDY00017558. We recorded data in sessions of approximately 20\,min length. In the rare event of magnetometer sensor drift during a session, we asked the human subject to recalibrate the devices and start over. % discarded the recording.
We recorded at least 4 sessions per human subject. Our training dataset consists of 970,493 data points collected from five human subjects.  %Our test dataset is completely separate from the training process. 
For the test dataset, the participants performed a series of common movements including crossing the arm on the chest and behind the head, waving, and walking in a figure-eight in a separate recording session. % Example poses are depicted in \Cref{fig:pred_examples}.
Further, we encouraged participants to perform movements typical for teleoperation, e.g., slow wrist movements in a straight line or orientation changes with a constant arm pose. % Like our training dataset, our test data was augmented with additional body orientations.

We augmented training and test data by retrospectively rotating the calibrated smartwatch orientation $\pmb \theta_\mathrm{sw}$ and smartphone body orientation~$\mathbf{r}_\mathrm{h}$ by a random angle around the up-axis. This augmentation artificially simulated additional body orientations. This was feasible because the remaining sensor measurements, such as the accelerometer or barometer, are in the watch's reference frame and therefore unaffected by global orientation changes. As a result of this augmentation process, our training dataset amounts to 4,259,746 data points and our test dataset amounts to 26,688 data points.

\subsection{Model Performance}

\begin{figure}[t]
\centerline{\includegraphics[width=\linewidth]{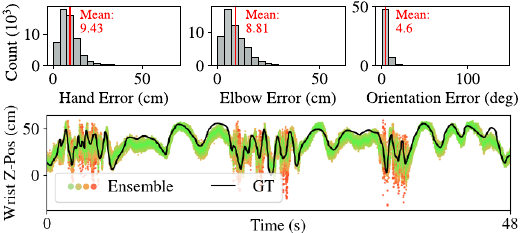}}
\caption{{\bf Top:} Prediction error distributions of DEnKF on the test dataset. {\bf Bottom:} An example for widening ensemble distributions, i.e., higher uncertainty, when the user moves fast. Ensemble members are colored according to their distance from the mean. GT is the Ground Truth.}
\label{fig:test_data_errors}
\end{figure}

To evaluate human pose estimation accuracy on the test dataset we employed Eucledian distance and angular difference metrics. To this end, we processed the DEnKF state ensemble outputs because they quantify the state uncertainty through the distribution of their ensemble members. %Depicted ensemble members in \Cref{fig:method,fig:test_data_errors}, demonstrate how movement speed and arm pose can affect the uncertainty. 
For example, in \Cref{fig:test_data_errors} the distributions widen when the wrist Z-position changes fast, indicating a lower confidence. To obtain a singular Euclidean distance to the ground truth, we averaged the rotation values of all ensemble members for the lower arm \gls{rot_larm}, upper arm \gls{rot_uarm}, and hip $\mathbf{q}_\mathrm{h}$. Assuming a default fixed lower arm length $l_\mathrm{l}$, fixed upper arm length $l_\mathrm{u}$, and fixed distance from hip to shoulder, we utilized forward kinematics to determine the relative XYZ coordinates of the wrist and elbow. This provided us with intuitive Euclidean distances for our evaluation and comparison with related work. \Cref{fig:test_data_errors} summarizes the performance of the DEnKF model on the test dataset. On average, hand positions were off by 9.43\,cm, elbow positions by 8.81\,cm, and body orientation by 4.6\,deg.

\begin{figure*}[ht]
\centering
\includegraphics[width=\linewidth]{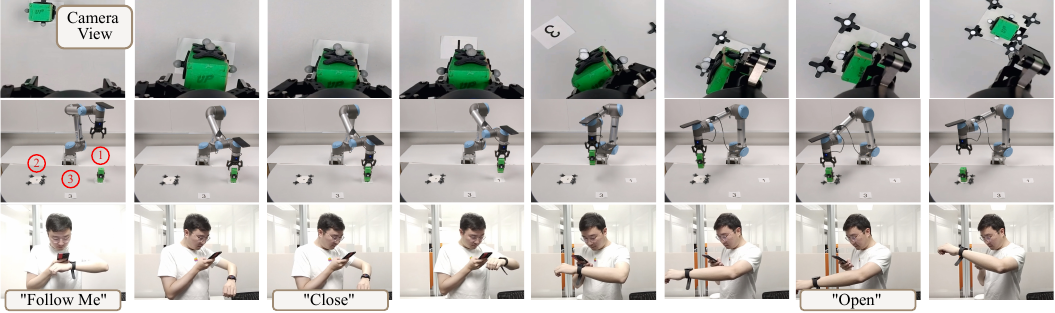}
\caption{An example for our teleoperation pick-and-place task. The task had 3 possible locations for cube and target. Human subjects watched the first-person camera view on the phone and controlled the robot through body movements and voice commands.}
\label{fig:pick_place}
\end{figure*}

\begin{table}[ht]
\begin{center}
\footnotesize
\caption{Model Comparison with related work}
\label{Tab:model_compare}
\scalebox{0.95}{
\begin{tabular}{c c c S[table-format=2.2] c c}
\toprule 
 &  & Free Forward-& \multicolumn{1}{c}{Wrist} & Elbow & Hip\\
Method & Anywhere & Facing Dir. & \multicolumn{1}{c}{(cm)} & (cm) & (deg) \\
\midrule
\cite{liu2022real} & $\times$ & $\checkmark$ & 10.93 & - & -\\
\cite{wei2021real} & $\checkmark$ & $\times$ & 8.50 & 8.50 & -\\
\cite{shen_i_2016} & $\checkmark$ & $\times$ & 9.20 & 7.90 & -\\
\cellcolor{gray!10} Ours & \cellcolor{gray!10}$\checkmark$ & \cellcolor{gray!10}$\checkmark$ & \cellcolor{gray!10}9.43 & \cellcolor{gray!10}8.81 & \cellcolor{gray!10}4.60\\
\bottomrule
\end{tabular}}
\end{center}
\end{table}

\Cref{Tab:model_compare} compares our results with other related works~\cite{wei2021real,liu2022real,shen_i_2016}. %Although the methods summarized in \Cref{Tab:model_compare} utilize distinct training data and models, the estimated wrist positions exhibit a similar average of approximately 9\,cm. 
The method of \cite{wei2021real} requires inference in the same environment where the training data was collected, therefore, it is not applicable \emph{anywhere}. Further, \cite{wei2021real} focuses on pose predictions for the wrist, omitting the rest of the body pose including elbow or hip orientation. Methods of \cite{liu2022real} and \cite{shen_i_2016} demonstrate lower errors for wrist and elbow but fix the user to a constant forward-facing direction. In contrast, iRoCo also provides an estimate of the Hip pose and allows for ubiquitous pose estimation regardless of location or changes in body orientation. Like all related works we compare to, iRoCo is real-time capable. Our framework achieves inference speed at $\sim$62\,Hz on a system using an Intel® Xeon(R) W-2125 CPU and NVIDIA GeForce RTX 2080 Ti.

\section{Real-Robot Applications}
% The iRoCo framework effectively addresses the need for a balanced integration of the flexibility offered by smart devices and the requirement for accurate control in robotics. Consequently, we introduce iRoCo, which combines DEnKF to establish a ubiquitous and versatile control interface. 
To further assess iRoCo's capabilities to facilitate ubiquitous robot control, we apply it in teleoperation and outdoor drone piloting tasks. 

\subsection{Teleoperation}

For the teleoperation task, participants remotely controlled a UR5 to pick-and-place a cuboid. We compared the task completion times and placement accuracy with iRoCo and OptiTrack as the baseline. % For tasks with iRoCo, the users were outside the laboratory removed from the UR5. For control with OptiTrack, the users had to be present in the laboratory to be within the trackable area of the cameras. 
Data collection was approved under the IRB of ASU under the ID STUDY00018521.

\textbf{Task Setup}: The first column of \Cref{fig:pick_place} details the task setup. The UR5 had operated on a tabletop surface with three marked locations. We placed a cuboid at one of these locations and at another a platform indicating the placement target. %The human subject then was instructed to control the UR5 to pick up the cube and place it on the target platform.
To control the UR5 remotely, the human subject wore a smartwatch and held a smartphone. We initialized a Zoom session between a smartphone mounted on the UR5 and the smartphone held by the human subject. This provided a first-person view. %while the user controlled the robot end-effector. %We marked three locations on the tabletop surface. %To control the end-effector, as depicted in \Cref{fig:coord_system}, we use the global hip rotation ${\bf r}_h$ to rotate a local right-handed coordinate system with the Y-axis up and Z-axis forward. %Using the state estimated from the smartwatch, we then extract the wrist position on the sagittal plane of the user in this reference frame. 
%The controlled UR5 matches its base joint to the oriented coordinate system and the end-effector to the user's local wrist position on the sagittal plane. 
Further, the smartwatch transcribed voice commands for gripper control. % We asked the human subject to complete six tasks with iRoCo and six with Optitrack.

\textbf{Task Procedure:} The entire \Cref{fig:pick_place} depicts the procedure of an example pick-and-place teleoperation task. In the beginning, the human subject started the iRoCo motion capture and then instructed the robot to follow their arm movements (``Follow Me''). This triggered the robot to follow the arm motions of the human subject as defined by our control modality (\Cref{subsec:control}). The subject then maneuvered the end-effector toward the cuboid and grabbed it by closing the gripper with a voice command (``Close''). Then, the subject lifted the end-effector and maneuvered it to the target platform. Once the cuboid hovered above the target platform, they opened the gripper with another voice command (``Open''). % We recorded the task completion time from the ``follow me'' until the ``open'' command as well as the cuboid placement distance from center of the target platform.

% \begin{wraptable}{l}{5.6cm}
% \vspace{-0.1in}
\begin{table}[h]
\begin{center}
\footnotesize
\caption{Pick-and-place task comparison}
\label{Tab:pick_place}
\begin{tabular}{ 
c
S[table-format=2.1] @{${}\pm{}$} S[table-format=2.1]
S[table-format=1.1] @{${}\pm{}$} S[table-format=1.1]
c
c
}
\toprule 
Method & \multicolumn{2}{c}{Time (s)} & \multicolumn{2}{c}{Dist. (cm)} &  \multicolumn{1}{c}{Success} & \multicolumn{1}{c}{Fail}\\
\midrule
Optitrack &59.8 & 16.5 & 4.5 & 2.9 & 29 & 1\\
iRoCo &71.3 & 25.6 & 6.3 & 6.5 & 28 & 2 \\
% \midrule
% Diff. & 13.6 & 28.9 & 1.8 & 6.7 & 1 & 1\\
\bottomrule
\end{tabular}
\end{center}
\end{table}
% \end{wraptable}

% Furthermore, a smartwatch interface is employed to control the arm's forward direction for two specific purposes: 1) Fine-tuning the lateral position of the end-effector to ensure precise alignment above the target object, and 2) Adjusting the vertical position of the end-effector for executing pick-and-place operations. 

% \begin{figure}
%     \centering
%     \includegraphics[width=\linewidth]{figures/shub_drone.pdf}
%     \caption{\textbf{Left:} The setup of the experiments, the targets are marked with an AprilTag and we confirm through the camera stream of the drone that it reached the target. \textbf{Right:} iRoCo uses the orientation to define the forward direction and the smartwatch pose to set drone elevation and distance. 
%     % This way, the user just has to point where the drone should fly to.
%     }
%     \label{fig:drone}
% \end{figure}

\begin{figure*}[ht]
\centering
\includegraphics[width=\linewidth]{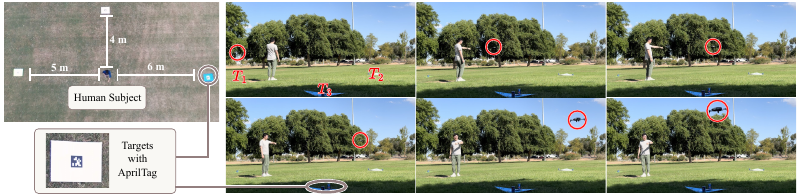}
\caption{\textbf{Left:} The drone piloting task setup. Targets ($T_1, T_2, T_3$) were marked with AprilTags~\cite{wang2016iros} to be detected through the camera stream of the drone. \textbf{Right:} Human subjects piloted the drone to all three targets in a given random order. This figure depicts the control with iRoCo, where the user only had to point toward the target and adjust elevation and distance with their wrist.}
%\vspace{-1\baselineskip}
\label{fig:drone_series}
\end{figure*}

\textbf{Results}: Five human subjects performed twelve pick-and-place tasks each, six with iRoCo and six with Optitrack. For every task, we altered the cuboid and target platform positions (1-2, 1-3, 2-1, \ldots). %Findings further confirm that iRoCo allows for effective control and manipulation of the UR5. 
\Cref{Tab:pick_place} summarizes the results. Three tasks failed because the UR5 knocked over the cuboid. If a task succeeded, we measured the time until task completion (``Follow Me'' until ``Open''). Further, we measured the distance from the placed cube to the center of the target location using OptiTrack. %Five participants completed the study, which accumulated 30 pick-and-place tasks with iRoCo and OptiTrack respectively. % Even though with iRoCo participants were outside of the laboratory and only used a smartwatch and smartphone to teleoperate the UR5, placement accuracy results are not significantly different to the baseline. However, task completion time was a bit slower.
On average, human subjects completed pick-and-place tasks $\sim$13\,s faster with OptiTrack than with iRoCo and placed the cuboid $\sim$1.8\,cm closer to the target. A paired t-test results in a \mbox{p-value} of 0.018 for completion times and 0.161 for distance measurements, indicating a significant difference in execution times but no significant difference in placement accuracy. This indicates that users might expect slightly longer task execution times when using iRoCo but they are able to achieve a similar level of placement accuracy for the task. %Therefore, results confirm iRoCo is effective for this teleoperation task. %We argue that this is acceptable considering 

%The clear strength of iRoCo is that it is ubiquitous. In contrast to wearing a retroreflective-marker suit and having to be in sight of a calibrated array of OptiTrack cameras, with iRoCo, the user can get their smartphone and watch, start wireless streaming, and control the UR5 from anywhere. 

% The users can effectively explore and navigate through UR5's workspace in order to perform tasks such as object localization and target manipulation. 
% This integration allows for effective control and manipulation of the UR5 arm using the human body orientation and wrist positions. 

% To assess performance, we measure two key metrics: the completion time and the 3D distance between the placed cube and the target location. The distance measurement was obtained using retroreflective markers attached to the object and target location. The timing measurement was taken from the moment the participant issued the ``follow me" voice command until the gripper fully opened and released the cube. Failure cases are identified when the user accidentally knocked over the cube with the gripper, rendering it impossible to pick up. We observed two failure cases with the wearable control and one failure case with the Optitrack control. Thus, our dataset consists of a total of 28 wearable control tasks and 29 Optitrack control tasks. \Cref{Tab:pick_place} concludes the comparison of using two systems.

\subsection{Drone Piloting}

In this drone piloting task, human subjects used iRoCo to fly a commercial Parrot Bebop 2 drone to three target locations. We compared iRoCo control to the original remote named SkyController as the baseline. % Intuitive drone control through body motions is attractive because it aids inexperienced users in becoming comfortable with drone piloting \cite{DroneTeleop}. 
this application further highlights the \emph{anywhere} aspect of iRoCo, as we conducted it outside and without any specialized infrastructure. The IRB of ASU approved data collection under the ID STUDY00018450.

\textbf{Task Setup:} The left part of \Cref{fig:drone_series} depicts an overview of our task setup. Human subjects stood in the middle of a field and had three targets around them. Targets were colored cardboard sheets marked with AprilTags~\cite{wang2016iros} placed at 4\,m, 5\,m, and 6\,m distance. The task was to hover the drone above these targets in a given random order. When the downward-facing camera stream of the drone recognized the AprilTag~ID, we considered the target reached and the drone had to fly to the next one.

\textbf{Task Procedure:} The right part of \Cref{fig:drone_series} depicts an example task procedure. We assigned the three targets in random order. In this example, the order was T1, T2, and then T3. Using iRoCo, the Human subject controlled the drone through their body orientation and wrist position. This way, motion capture control with iRoCo enabled the subject to simply orient themselves toward a target and adjust with wrist movements while keeping the target and the drone in view. In the bottom right picture of \Cref{fig:drone_series} the drone hovers above the final target and the task is completed.  %The drone uses GPS and internal IMUs to follow global positions from iRoCo in a stable trajectory. %as it was with the UR5 end-effector in the previous application. We place the user in the middle of the targets, such that they have to turn their body to reach all of them. 
% An example flight is depicted in \Cref{fig:drone_series}. We assign the three targets in random order. Then, the participant completes two runs, one with iRoCo and one with the original remote controller. %The controller has two joysticks, one for elevation and roll and one for forward, back, left, and right.

\textbf{Results:} %Collected data confirms that iRoCo facilitates more intuitive drone control. 
Ten human subjects completed the task with both iRoCo and with the SkyController. We measured the task completion time as the time from reaching the first target until the last. Post-completion, subjects answered a subjective task load index questionnaire (NASA-TLX)\cite{feick2020virtual}. \Cref{Tab:piloting} summarizes our findings. On average, participants completed the task $\sim$19\,s faster with iRoCo than with the original remote. This seems reasonable because motion capture allowed to keep the drone in view, while, with the remote, subjects tended to alternate between checking the display and looking at the drone to keep a mental map of where the drone is in relation to the targets. 
We highlight three categories of the NASA-TLX questionnaires, which support this assessment. Subjects reported a lower mental load with iRoCo and a considerably lower frustration score with a low standard deviation. However, the iRoCo system scored higher on the physical load index because the subjects had to control with turns and arm movements.

% Altogether, both applications demonstrate that iRoCo is a viable alternative to established teleoperation and drone piloting modes. %The ubiquitous and versatile applications of iRoCo are strong because the user can rely on familiar smartwatches and smartphones for intuitive robot control. 
% The possibility to use common smartdevices emphasizes the "anytime anywhere" aspect of iRoCo and promises to 

\begin{table}[h]
% \vspace{-0.1in}
\begin{center}
\caption{Piloting comparison (10 Human Subjects)}
\label{Tab:piloting}
\begin{tabular}{ 
c
S[table-format=2.1] @{${}\pm{}$} S[table-format=2.1]
S[table-format=1.1] @{${}\pm{}$} S[table-format=1.1]
S[table-format=1.1] @{${}\pm{}$} S[table-format=1.1]
S[table-format=1.1] @{${}\pm{}$} S[table-format=1.1]
}
\toprule 
 & 
\multicolumn{2}{c}{Objective}  & 
\multicolumn{6}{c}{Subjective Task Load Index}\\
\cmidrule(lr){2-3}
\cmidrule(lr){4-9} 
Method & 
\multicolumn{2}{c}{Time (s)} & 
\multicolumn{2}{c}{Mental} & 
\multicolumn{2}{c}{Physical} &
\multicolumn{2}{c}{Frustration} \\
\midrule
Remote&59.7 & 27.8 & 7.3 & 3.6 & 3.7 & 2.9 & 4.5 & 2.4 \\
iRoCo &40.5 & 10.1 & 4.7 & 4.4 & 6.1 & 4.3 & 2.0 & 1.5 \\
\bottomrule
\end{tabular}
\end{center}
\end{table}

\section{Conclusion}
This work introduced iRoCo, an intuitive and ubiquitous system to control robots with a smartwatch and smartphone. By incorporating a differentiable ensemble Kalman filter algorithm, it optimizes the balance between motion capture accuracy and unconstrained user movement. We demonstrated its effectiveness in practical teleoperation and drone piloting tasks with time and distance metrics as well as subjective task load metrics. Results confirm that, with only a smartwatch and a smartphone, iRoCo offers new and intriguing possibilities for ubiquitous robot control and human-robot collaboration.

\bibliographystyle{IEEEtran}
\scriptsize{
\bibliography{references}

\begin{thebibliography}{10}
\providecommand{\url}[1]{#1}
\csname url@rmstyle\endcsname
\providecommand{\newblock}{\relax}
\providecommand{\bibinfo}[2]{#2}
\providecommand\BIBentrySTDinterwordspacing{\spaceskip=0pt\relax}
\providecommand\BIBentryALTinterwordstretchfactor{4}
\providecommand\BIBentryALTinterwordspacing{\spaceskip=\fontdimen2\font plus
\BIBentryALTinterwordstretchfactor\fontdimen3\font minus
  \fontdimen4\font\relax}
\providecommand\BIBforeignlanguage[2]{{%
\expandafter\ifx\csname l@#1\endcsname\relax
\typeout{** WARNING: IEEEtran.bst: No hyphenation pattern has been}%
\typeout{** loaded for the language `#1'. Using the pattern for}%
\typeout{** the default language instead.}%
\else
\language=\csname l@#1\endcsname
\fi
#2}}

\bibitem{ajoudani2018progress}
A.~Ajoudani, A.~M. Zanchettin, S.~Ivaldi, A.~Albu-Sch{\"a}ffer, K.~Kosuge, and
  O.~Khatib, ``Progress and prospects of the human--robot collaboration,''
  \emph{Autonomous Robots}, vol.~42, pp. 957--975, 2018.

\bibitem{8968007}
I.~Patzer and T.~Asfour, ``Minimal sensor setup in lower limb exoskeletons for
  motion classification based on multi-modal sensor data,'' in \emph{2019
  IEEE/RSJ International Conference on Intelligent Robots and Systems (IROS)},
  2019, pp. 8164--8170.

\bibitem{NIPS1996_68d13cf2}
S.~Schaal, ``Learning from demonstration,'' in \emph{Advances in Neural
  Information Processing Systems}, M.~Mozer, M.~Jordan, and T.~Petsche, Eds.,
  vol.~9.\hskip 1em plus 0.5em minus 0.4em\relax MIT Press, 1996.

\bibitem{BenAmorNKKP2014}
H.~Ben~Amor, G.~Neumann, S.~Kamthe, O.~Kroemer, and J.~Peters, ``Interaction
  primitives for human-robot cooperation tasks,'' in \emph{Proceedings of 2014
  IEEE International Conference on Robotics and Automation}.\hskip 1em plus
  0.5em minus 0.4em\relax IEEE, 2014, pp. 2831--2837.

\bibitem{nagymate_application_2018}
G.~Nagymáté and R.~M.~Kiss, ``Application of optitrack motion capture systems
  in human movement analysis: A systematic literature review,'' \emph{Recent
  Innovations in Mechatronics}, vol.~5, no.~1., p. 1–9., Jul. 2018.

\bibitem{DESMARAIS2021103275}
Y.~Desmarais, D.~Mottet, P.~Slangen, and P.~Montesinos, ``A review of 3d human
  pose estimation algorithms for markerless motion capture,'' \emph{Computer
  Vision and Image Understanding}, vol. 212, p. 103275, 2021.

\bibitem{DBLP:journals/corr/MarcardRBP17}
T.~von Marcard, B.~Rosenhahn, M.~J. Black, and G.~Pons{-}Moll, ``Sparse
  inertial poser: Automatic 3d human pose estimation from sparse imus,''
  \emph{CoRR}, vol. abs/1703.08014, 2017.

\bibitem{yi2021transpose}
X.~Yi, Y.~Zhou, and F.~Xu, ``Transpose: Real-time 3d human translation and pose
  estimation with six inertial sensors,'' \emph{ACM Transactions on Graphics
  (TOG)}, vol.~40, no.~4, pp. 1--13, 2021.

\bibitem{roetenberg2009xsens}
D.~Roetenberg, H.~Luinge, P.~Slycke, \emph{et~al.}, ``Xsens mvn: Full 6dof
  human motion tracking using miniature inertial sensors,'' \emph{Xsens Motion
  Technologies BV, Tech. Rep}, vol.~1, pp. 1--7, 2009.

\bibitem{villani2020humans}
V.~Villani, B.~Capelli, C.~Secchi, C.~Fantuzzi, and L.~Sabattini, ``Humans
  interacting with multi-robot systems: a natural affect-based approach,''
  \emph{Autonomous Robots}, vol.~44, pp. 601--616, 2020.

\bibitem{weigend2023anytime}
F.~C. Weigend, S.~Sonawani, M.~Drolet, and H.~B. Amor, ``Anytime, anywhere:
  Human arm pose from smartwatch data for ubiquitous robot control and
  teleoperation,'' in \emph{2023 IEEE/RSJ International Conference on
  Intelligent Robots and Systems (IROS)}, 2023, pp. 3811--3818.

\bibitem{shen_i_2016}
S.~Shen, H.~Wang, and R.~Roy~Choudhury, ``I am a smartwatch and i can track my
  user's arm,'' in \emph{Proceedings of the 14th Annual International
  Conference on Mobile Systems, Applications, and Services}.\hskip 1em plus
  0.5em minus 0.4em\relax {ACM}, 2016, pp. 85--96.

\bibitem{wei2021real}
W.~Wei, K.~Kurita, J.~Kuang, and A.~Gao, ``Real-time limb motion tracking with
  a single imu sensor for physical therapy exercises,'' in \emph{2021 43rd
  Annual International Conference of the IEEE Engineering in Medicine \&
  Biology Society (EMBC)}.\hskip 1em plus 0.5em minus 0.4em\relax IEEE, 2021,
  pp. 7152--7157.

\bibitem{liu2022real}
M.~Liu, S.~Yang, W.~Chomsin, and W.~Du, ``Real-time tracking of smartwatch
  orientation and location by multitask learning,'' in \emph{Proceedings of the
  20th ACM Conference on Embedded Networked Sensor Systems}, 2022, pp.
  120--133.

\bibitem{villani2017interacting}
V.~Villani, L.~Sabattini, G.~Riggio, A.~Levratti, C.~Secchi, and C.~Fantuzzi,
  ``Interacting with a mobile robot with a natural infrastructure-less
  interface,'' \emph{IFAC-PapersOnLine}, vol.~50, no.~1, pp. 12\,753--12\,758,
  2017.

\bibitem{kloss2021train}
A.~Kloss, G.~Martius, and J.~Bohg, ``How to train your differentiable filter,''
  \emph{Autonomous Robots}, pp. 1--18, 2021.

\bibitem{liu2023enhancing}
X.~Liu, G.~Clark, J.~Campbell, Y.~Zhou, and H.~B. Amor, ``Enhancing state
  estimation in robots: A data-driven approach with differentiable ensemble
  kalman filters,'' in \emph{2023 IEEE/RSJ International Conference on
  Intelligent Robots and Systems (IROS)}, 2023, pp. 1947--1954.

\bibitem{liu2023alpha}
X.~Liu, Y.~Zhou, S.~Ikemoto, and H.~B. Amor, ``$\alpha$-mdf: An attention-based
  multimodal differentiable filter for robot state estimation,'' in \emph{7th
  Annual Conference on Robot Learning}, 2023.

\bibitem{lee2015smartwatch}
B.-G. Lee, B.-L. Lee, and W.-Y. Chung, ``Smartwatch-based driver alertness
  monitoring with wearable motion and physiological sensor,'' in \emph{2015
  37th Annual International Conference of the IEEE Engineering in Medicine and
  Biology Society (EMBC)}.\hskip 1em plus 0.5em minus 0.4em\relax IEEE, 2015,
  pp. 6126--6129.

\bibitem{wicaksono2022towards}
H.~Wicaksono, I.~Sugiarto, P.~Santoso, G.~Ricardo, and J.~Halim, ``Towards
  autonomous robot application and human pose detection for elders
  monitoring,'' in \emph{2022 6th International Conference on Information
  Technology, Information Systems and Electrical Engineering (ICITISEE)}.\hskip
  1em plus 0.5em minus 0.4em\relax IEEE, 2022, pp. 772--776.

\bibitem{NEURIPS2018_5cf68969}
S.~S. Rangapuram, M.~W. Seeger, J.~Gasthaus, L.~Stella, Y.~Wang, and
  T.~Januschowski, ``Deep state space models for time series forecasting,'' in
  \emph{Advances in Neural Information Processing Systems}, S.~Bengio,
  H.~Wallach, H.~Larochelle, K.~Grauman, N.~Cesa-Bianchi, and R.~Garnett, Eds.,
  vol.~31.\hskip 1em plus 0.5em minus 0.4em\relax Curran Associates, Inc.,
  2018.

\bibitem{klushyn2021latent}
A.~Klushyn, R.~Kurle, M.~Soelch, B.~Cseke, and P.~van~der Smagt, ``Latent
  matters: Learning deep state-space models,'' \emph{Advances in Neural
  Information Processing Systems}, vol.~34, 2021.

\bibitem{wagstaff2022self}
B.~Wagstaff, E.~Wise, and J.~Kelly, ``A self-supervised, differentiable kalman
  filter for uncertainty-aware visual-inertial odometry,'' in \emph{2022
  IEEE/ASME International Conference on Advanced Intelligent Mechatronics
  (AIM)}.\hskip 1em plus 0.5em minus 0.4em\relax IEEE, 2022, pp. 1388--1395.

\bibitem{lee2020multimodal}
M.~A. Lee, B.~Yi, R.~Mart{\'\i}n-Mart{\'\i}n, S.~Savarese, and J.~Bohg,
  ``Multimodal sensor fusion with differentiable filters,'' in \emph{2020
  IEEE/RSJ International Conference on Intelligent Robots and Systems
  (IROS)}.\hskip 1em plus 0.5em minus 0.4em\relax IEEE, 2020, pp.
  10\,444--10\,451.

\bibitem{liu2023learning}
X.~Liu, S.~Ikemoto, Y.~Yoshimitsu, and H.~B. Amor, ``Learning soft robot
  dynamics using differentiable kalman filters and spatio-temporal
  embeddings,'' in \emph{2023 IEEE/RSJ International Conference on Intelligent
  Robots and Systems (IROS)}, 2023, pp. 2550--2557.

\bibitem{zhou_continuity_2019}
Y.~Zhou, C.~Barnes, J.~Lu, J.~Yang, and H.~Li, ``On the continuity of rotation
  representations in neural networks,'' in \emph{2019 {IEEE}/{CVF} Conference
  on Computer Vision and Pattern Recognition ({CVPR})}.\hskip 1em plus 0.5em
  minus 0.4em\relax {IEEE}, 2019, pp. 5738--5746.

\bibitem{evensen2003ensemble}
G.~Evensen, ``The ensemble kalman filter: Theoretical formulation and practical
  implementation,'' \emph{Ocean dynamics}, vol.~53, no.~4, pp. 343--367, 2003.

\bibitem{gal2016dropout}
Y.~Gal and Z.~Ghahramani, ``Dropout as a bayesian approximation: Representing
  model uncertainty in deep learning,'' in \emph{international conference on
  machine learning}.\hskip 1em plus 0.5em minus 0.4em\relax PMLR, 2016, pp.
  1050--1059.

\bibitem{wang2016iros}
J.~Wang and E.~Olson, ``{AprilTag} 2: Efficient and robust fiducial
  detection,'' in \emph{Proceedings of the {IEEE/RSJ} International Conference
  on Intelligent Robots and Systems {(IROS)}}, October 2016.

\bibitem{feick2020virtual}
M.~Feick, N.~Kleer, A.~Tang, and A.~Kr{\"u}ger, ``The virtual reality
  questionnaire toolkit,'' in \emph{Adjunct Proceedings of the 33rd Annual ACM
  Symposium on User Interface Software and Technology}, 2020, pp. 68--69.

\end{thebibliography}
}

\end{document}